# Social Computational Design Method for Generating Product Shapes with GAN and Transformer Models

Maolin Yang*. Pingyu Jiang*

*State Key Laboratory for Manufacturing Systems Engineering, Xi'an Jiaotong University

Xi'an 710049, China (Tel: +86-29-83395396; e-mail: {maolin, pjiang}@mail.xjtu.edu.cn)

**Abstract**: A social computational design method is established, aiming at taking advantages of the fast-developing artificial intelligence technologies for intelligent product design. Supported with multi-agent system, shape grammar, Generative adversarial network, Bayesian network, Transformer, etc., the method is able to define the design solution space, prepare training samples, and eventually acquire an intelligent model that can recommend design solutions according to incomplete solutions for given design tasks. Product shape design is used as entry point to demonstrate the method, however, the method can be applied to tasks rather than shape design when the solutions can be properly coded.

*Keywords*: social computational design; generative design; computational design; product shape design

## 1. INTRODUCTION

The fast development of sharing economy, internet technologies, artificial intelligence algorithms, and the computing power of personal computers have boosted the emergence of a new kind of social computational design method (Yang & Jiang, 2019). The most prominent characteristics of the social computational design method compared to traditional product design method is that large numbers of socialized designers with different background participated in the same design task through online collaboration tools, and by applying advanced intelligence algorithms such as *Generative adversarial network* (GAN), *Variational auto-encoder* (Yoo et al., 2021) and *Transformer* (Ganin et al., 2021), etc., large numbers of design solutions for the design task can be efficiently generated not only based on the design ideas of the designers but also historical design data (as shown in Fig. 1.)

Social computational design method has gained the attention from both academic and industrial field. **For academic researches**, Val and Muiños-Landin (2020) proposed a variational autoencoder enabled approach to support artificial creativity augmentation among socialized designers. Yang et al. (2021) proposed a multi-layer product design method enabled with Blackboard control system, and the method is able to utilize the collective intelligence from socialized designers for product innovation. Ganin et al.(2021) proposed a Transformer based sketch design method where large number of sketches from socialized designers are used as training data. **For industrial applications**, *Autodesk* developed a generative design software module that can generate large numbers of alternative design solutions according to the initial geometry and constraints (Vlah et al., 2020). Based on this technique, *Airbus* utilized generative design approach to design the structure of the gate of the galley on Airbus A320 (Airbus, 2021).

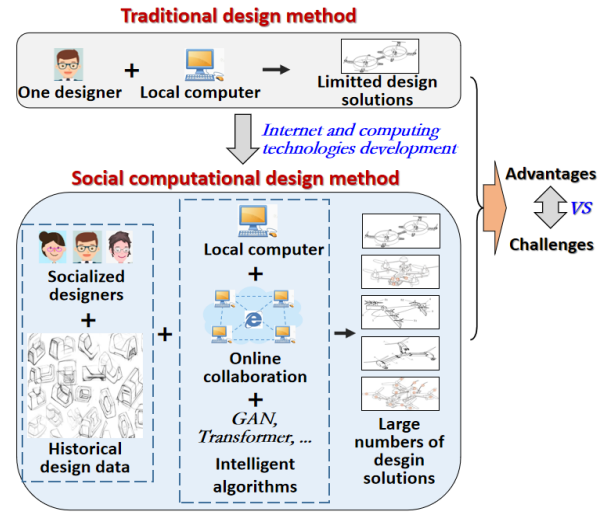

Fig. 1. The changes emerged in product design method resulted from the development of internet technologies, artificial intelligence algorithms, computing power, etc., together with the advantages and challenges that comes along with.

Social computational design method has many **advantages**, such as large numbers of targeted design solutions with relative lower cost, strong design inspiration resulted from the collective intelligence of socialized designers, and more importantly the design solution space could be thoroughly searched with computers and algorithms. However, there are also **challenges** need to be considered before making full use of the aforementioned advantages. For example, how to organize the efficiently online collaboration among the socialized designers from different background, how to establish an unified expression method of the design solutions from different designers, how exactly to customize and apply the intelligent algorithms for design supporting task, how to acquire sufficient training data for the supervised intelligent algorithms, etc.

To address the challenges mentioned above, a social computational design method is established, and product shape design is used as entry point to demonstrate its operability.

2. THE GENERAL ROAD MAP OF THE METHOD

The general road map of social computational design method for generating product shapes can be separated into two parts, which are key enabling technologies (KETs) and application implementation method, as shown in Fig. 2.

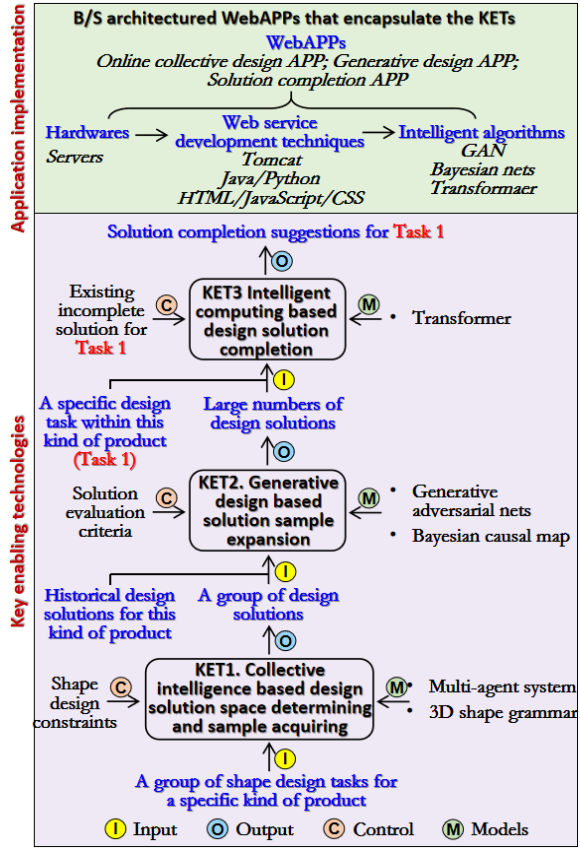

Fig. 2. The general road map of the social computational design method for generating product shapes

The KETs, which contains three main technologies, describe a complete method to conduct social computational design from training data preparation to final design solution completion. The relationships among the three KETs are described with IDEF0 chart, as shown in the lower part of Fig.2. Firstly KET1 is applied to generate a group of design solutions for a group of shape design tasks for a specific kind of product based on the collective intelligence of socialized designers. Secondly, the solutions and other historical design solutions for this kind of product can be used to generate large numbers of design solutions through KET2. Thirdly, the large number of solutions generated through KET2 can be used to train a design solution completion suggestion model through KET3, and the trained model is able to complete any incomplete solutions for a new design task within the range of this kind of product (denoted as *Task 1* in Fig.2). More detailed description of each KET are described in Section 3.

The application implementation method is for supporting the application of the KETs through a group of B/S architectured WebAPPs, which include at least three WebAPPs that encapsulate each of the the three KETs, as shown in the upper part of Fig.2. The basis that support the development and operation of the WebAPPs include servers, web service development techniques, and customized intelligent algorithms. An example of the application implementation is discussed in Section 4.

3. THE KEY ENABLING TECHNOLOGIES

This section describes the implementation method of each KET.

*3.1 Collective intelligence based design solution space and sample node acquiring*

The purpose of KET1 is to define the solution space for a specific kind of product shape design task, and acquire a set of sample nodes in the space that contain the collective intelligence of socialized designers. Due to the characteristics of socialized designers, which are self-driven, self-organized, with different social backgrounds, etc., the design activities participated by them are usually in the form of asynchronous, distributed, and stranger online interaction. Therefore KET1 has to be able to support collaborative design process management and at the same time considering the characteristics of social interaction. KET1 can be separated into three parts, and the relationships among them are shown in Fig.3.

1) Establishment of the 3D shape grammar for the target product. The application of 3D shape grammar has mainly two purpose. First, providing an unified language for the designers to collaborate on the same task. Second, it is easier to transform the design solutions expressed with grammar rules into the training data for KET2 and KET3. Establishment of the grammar for the target product can be achieved through three steps, which are feature engineering based shape units determining, parameterized 3D shape building rules determining, and determining the constraints for the rules, as shown in Fig.3 (1), where the grammar for the shape design of Drone is used as example.

2) Multi-agent system based social design management. The multi-agent system here is to support the orderly collaboration among different roles of participators, as shown in Fig.3 (2). After a *Task publisher* publishes a shape design task through its corresponding *Task management agent*, socialized designers can check the progress of the task recorded in the *Design process recording & control agent* through its corresponding *Design agent*, and submit his partial or complete solutions if he is able and willing to participate. During the process, *Shape grammar engineer* would upload predefined shape grammar to a *Shape grammar management agent*, and the designers can check the grammar by interacting with the *Shape grammar management agent*. Besides, if a solution submitted by a designer did not follow the constraints defined by the grammar, the *Shape grammar management agent* would send an error information to the corresponding designer. Eventually, the *Task publisher* can acquire all the submitted solutions from the *Design process recording &*

*control agent*, and the contribution degrees of all the participators of a specific design solution can be estimated automatically by a *Designer contribution degree estimation agent* (our previous work established an estimation method (Yang et al., 2021)).

3) Multi-dimension vector design space representation. Here multi-dimension vector space is used to represent the design solution space and the solutions in it. After the shape grammar for a specific kind of product has been established, the space of the design solutions that can be generated with the grammar is determined. Each solution can be abstracted as a node in the space, and the grammar rule sequence that generates a solution can be abstracted as the vectors whose summation starts from the origin node of the space and ends at the solution node.

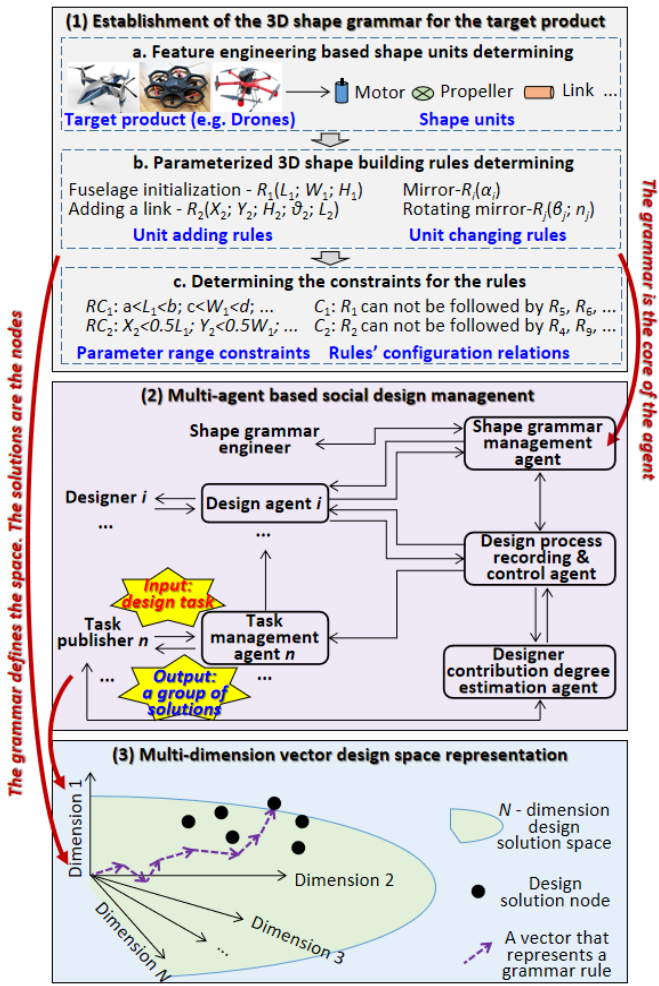

Fig. 3. The technology road map for collective intelligence based design solution space and sample node acquiring (KET1)

### 3.2 Generative design based solution sample expansion

KET2 has mainly two purposes. First, generating large numbers of alternative design solutions for a given shape design task on the basis of the initial solutions generated through KET1. Second, preparing the training data for KET3 (i.e. data enhancement). The implementation process of KET2 can be separated into three steps, as shown in Fig.4.

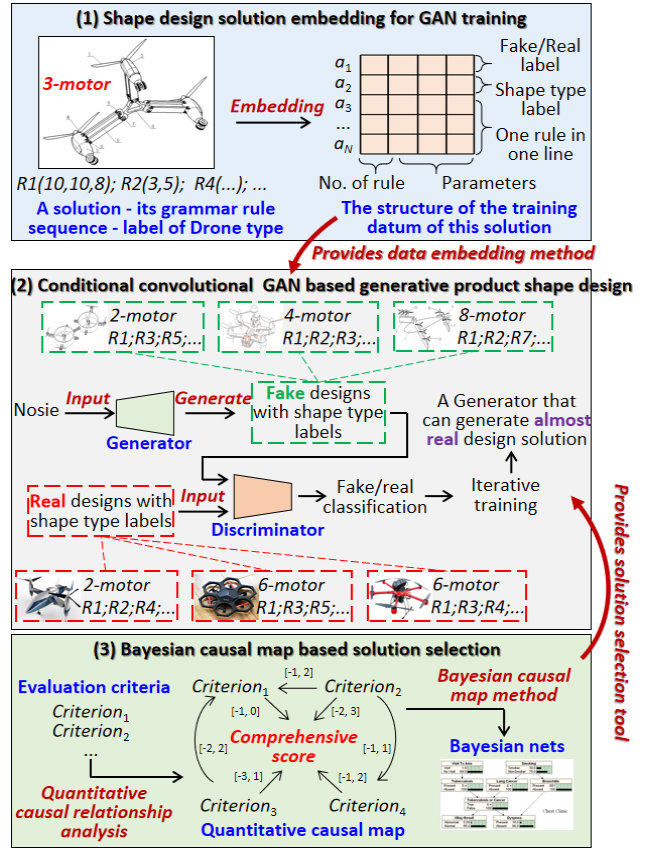

Fig. 4. The technology road map for generative design based design solution sample expansion (KET2)

1) Shape design solution embedding for GAN training. In KET1, all the shape design solutions are represented in the form of parameterized 3D shape grammar rules. These design solutions have to be embedded before used as the training data for GAN model. Here we suggest conditional convolutional GAN, and its suitable training data structure is illustrated in Fig. 4 (1). It is worth mention that many GAN based generative design models have been proposed in the past a few years (Regenwetter et al., 2021), and the reason to use conditional GAN is for the capability of generating design solutions of different shape design types (e.g. 4-motor Drone, 2-motor Drone), and the reason to use convolutional GAN is for convenient solution embedding.

2) Conditional convolutional GAN based generative product shape design. As illustrated in Fig.4 (2), real designs with their shape type labels are used to train the GAN model. With sufficient training data and proper training, eventually the trained Generator would be able to generate almost real design solutions together with their shape type labels (in the form of data structure in KET2(1)) from noise input, and the Discriminator would not be able to discriminate these almost real solutions from the real ones. The mechanism behind the training process is that the GAN model learns the shape unit configuration pattern from the training data, and then generate almost real configuration of the shape units within the design solution space defined by the established shape grammar. It is worth mention that historical real design solutions can be combined with the solutions acquired from KET1 for the training, as long as they are related to the design task (in this case, as long as they are drones).

3) Bayesian causal map based solution selection. The solutions generated with GAN would inevitably have flaws, and therefore have to pass an evaluation and selection procedure. Here we suggest using Bayesian causal map, which is a kind of Bayesian network transformed from quantitative causal map (Yang & Jiang, 2020). The reason to use Bayesian network for solution evaluation is because of its convenience when processing incomplete evaluation inputs. For example, if there were ten criteria modeled with the Bayesian network, and only seven (or even less) of the criteria can be provided with evaluation inputs for a target solution, the Bayesian network can still generate a comprehensive score for the target solution based on the evaluation inputs. The reason to use Bayesian causal map approach to build the Bayesian network is because it is an operable method when there are not sufficient historical causal data.

*3.3 Intelligent computing based design solution completion*

Transformer model was developed on the basis of Recurrent neural network and seq2seq models, and it is good at learning the pattern of sequence data (Ribeiro et al., 2020). Here, Transformer is applied to predict the rest shape grammar rule sequence for an incomplete solution according to the rule sequence that represents the incomplete solution. The entire approach can be separated into three steps, as shown in Fig. 5.

1) Designing the data structure for Transformer training. Different form the training data for the GAN model where each design solution and its shape type label are directly used for training after embedding, each piece of training data for the Transformer model need to be further separated into two parts, which are the input part and output part, and there are no Fake/Real label in the training data for Transformer, as shown in Fig.5(1). It should be noted that different separating strategies could be used for different training scenarios. For example, the data structure in Fig.5(1) is for the scenario of predicting the later $M$-$m$ pieces of shape grammar rules according to the earlier $m$ pieces of rules.

2) Training the Transformer model for grammar rule sequence prediction. This step is about training a Transformer model that can predict the next a few possible rules for an incomplete design solution. As illustrated in Fig. 5 (2), large amount of shape design solutions are transformed into training data with the methods in KET1 (1), KET2 (1), and KET3 (1), successively. Here we suggest using the almost real solutions generated with the GAN model to prepare the training data (i.e. the GAN model is used as a data enforcement method for Transformer training). After proper training, the Transformer model would be able to predict a sequence of shape grammar rules that match with the sequence of rules that represent an incomplete shape design solution. In another word, the predicted rule sequence can be attached to the rule sequence of the incomplete solution, and together the two sequences represent a complete design solution.

3) Transformer based grammar rule sequence recommendation. As illustrated in Fig. 5 (3), after the Transformer model has been trained, given a new shape design task and an incomplete design solution for this task in the form of shape grammar rules, the trained Transformer model can recommend a group of rule sequences, each of which can be attached to the already-have rules to represent a meaningful product shape. The designer can select from the recommendations and maybe make modifications on them. After repeating the human-in-loop design and intelligent recommendation a few rounds, the resign task would be completed.

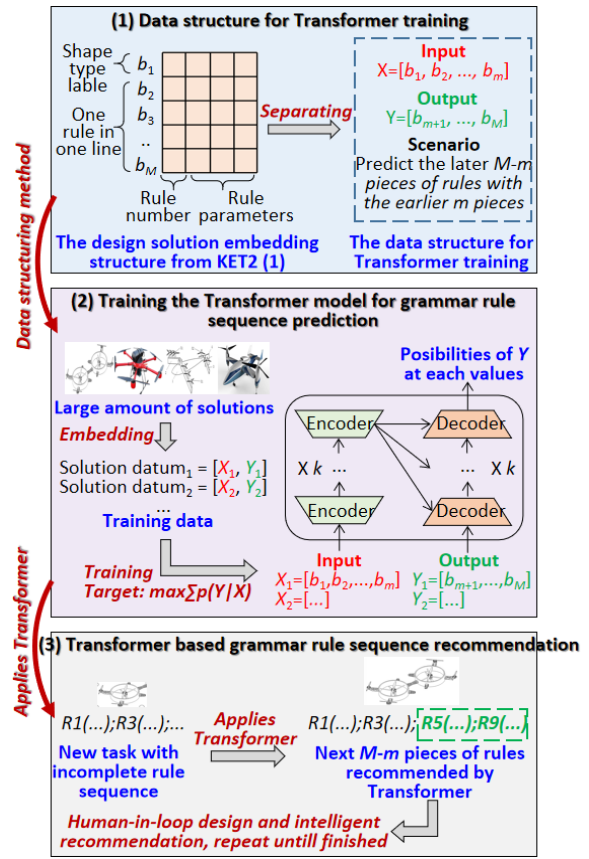

Fig. 5. The technology road map for intelligent computing based design solution completion (KET3)

## 4. CASE STUDY

Here, a case study of Drone shape design is used as example to introduce the approach of applying the methods in Section 2 and 3. The entire approach can be implemented with three steps.

1) Developing a social computational design platform. The basic functions of the platform and its main interfaces are shown in Fig.6 (1), and note that the platform should have a group of design supporting WebAPPs (bottom right of Fig.6 (1)).

2) Developing the essential social computational design supporting WebAPPs encoded with the KETs. Here we suggest developing three independent WebAPPs, as shown in Fig.6 (2).

3) Using the WebAPPs to support social computational design tasks, as illustrated in Fig.6 (3).

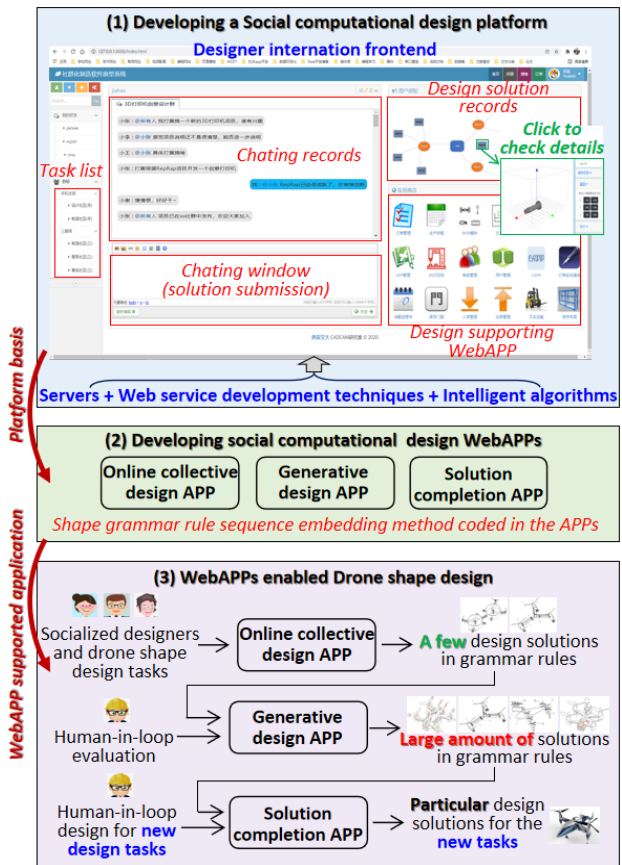

Fig. 6. The approach of applying the social computational design method for Drone shape design

## 5. DISCUSSION AND CONCLUSION

A social computational design method for generating product shapes together with its KETs is discussed in this paper, aiming at providing an operable method to take advantages of the fast developing internet technologies, intelligent algorithms, and computing power for product shape design.

The entire method can be considered as the combination of manual feature engineering (where shape grammar units are manually established by identifying most optimal function & structure features in product shape) and data driven pattern learning (where intelligent computing models are trained to learn the feature configuration patterns and feature sequence patterns from the training data of shape design solutions).

KET1 is mainly for establishing the multi-dimension vector design space of a specific kind of product, and acquiring design solution sample nodes in the space. After establishing the shape grammar for the kind of product, the design solution space have been determined. Each design solution is a sample node in this design space, and the grammar rule sequence that builds up this solution can be abstracted as the vectors whose summation vector starts from the origin node of the space and ends at the solution node.

KET2 is for identifying all the possible routes from the origin node to a meaningful design solution node within the design space according to the **vector configuration patterns** it learned from the training data, and in this way it can generate a large enough data set for KET3 training (i.e. data enforcement). It is worth mention that the vector configuration patterns are actually the grammar rule configuration patterns, and more deeply the structure/function feature configuration patterns of this kind of product. This is why KET1 is important, where real design solutions are generated based on the collective intelligence of multiple socialized designers. Because only when the GAN model learns from different designers, it can learn more patterns of feature configuration, and consequently become more "creative" after been trained (each designer would have his own design habits which result in a group of similar patterns).

After acquiring a large enough data set, KET3 is to learn the **sequence pattern among the vectors** that lead to each solution node in the design solution space. In another word, it learns which vector can or should be added to a previous vector give the vectors that already exist in this route, and this is the mechanism behind the Transformer based design solution completion. It is also worth mention that KET1 and 2 can be applied independently without KET3 if the purpose is only to acquire large numbers of alternatives for a group of manually generated solutions. However, KET 1 and 2 are the must for KET3, as it is difficult to acquire enough training data otherwise.

The main contribution of this paper is that it provides a complete and operable method for social computational design from data preparation to model application, and the method can be migrated to other product design domains when the design solutions can be coded into data structure suitable for the intelligent computing models.

## 6. ACKNOWLEDGEMENT

The work in this paper is supported by National Natural Science Foundation of China (No. 51975464).